\documentclass[11pt,a4paper]{article}
\usepackage[hyperref]{eacl2021}
\usepackage{times}
\usepackage{latexsym}

\usepackage{times}
\usepackage{url}
\usepackage{latexsym}
\usepackage{url}
\usepackage{hyperref}
\usepackage{booktabs}
\usepackage{multirow}
\usepackage{float}

\usepackage{array}
\usepackage{makecell}
\usepackage{graphicx}

\usepackage{microtype}
\usepackage{color,soul}
\definecolor{pink}{rgb}{1,0.8,0.8}
\definecolor{ltgreen}{rgb}{0.7,1,0.7}
\definecolor{ltblue}{rgb}{0.7,0.8,1}

\usepackage[prependcaption,textsize=scriptsize]{todonotes}
\aclfinalcopy % Uncomment this line for the final submission
\def\aclpaperid{1197} %  Enter the acl Paper ID here

\title{Facilitating Terminology Translation with Target Lemma Annotations}

\author{Toms Bergmanis$^{\dagger\ddagger}$ \and Mārcis Pinnis$^{\dagger\ddagger}$  \\ \\
  $^\dagger$Tilde / Vienības gatve 75A, Riga, Latvia \\
  $^\ddagger$Faculty of Computing, University of Latvia / Raiņa bulv.  19, Riga, Latvia\\ 
  {\tt \{firstname.lastname\}@tilde.lv}
}

\date{}

\begin{document}
\maketitle
\begin{abstract}
Most of the recent work on terminology integration in machine translation has assumed that terminology translations are given already inflected in forms that are suitable for the target language sentence. In day-to-day work of professional translators, however, it is seldom the case as translators work with bilingual glossaries where terms are given in their dictionary forms; finding the right target language form is part of the translation process. We argue that the requirement for apriori specified target language forms is unrealistic and impedes the practical applicability of previous work. In this work, we propose to train machine translation systems using a source-side data augmentation method\footnote{Relevant materials and code: \url{https://github.com/tilde-nlp/terminology_translation}} that annotates randomly selected source language words with their target language lemmas. We show that systems trained on such augmented data are readily usable for terminology integration in real-life translation scenarios. 
Our experiments on terminology translation into the morphologically complex Baltic and Uralic languages show an improvement of up to 7 BLEU points over baseline systems with no means for terminology integration and an average improvement of 4 BLEU points over the previous work.
Results of the human evaluation indicate a 47.7\% absolute improvement over the previous work in term translation accuracy when translating into Latvian. 

\end{abstract}

\section{Introduction}
Translation into morphologically complex languages involves 1) making a lexical choice for a word in the target language and 2) finding its morphological form that is suitable for the morpho-syntactic context of the target sentence. Most of the recent work on terminology translation, however, has assumed that the correct morphological forms are apriori known \cite{hokamp-liu-2017-lexically,post2018fast,hasler-etal-2018-neural,dinu-etal-2019-training,song2020alignment,susanto2020lexically,dougal-lonsdale-2020-improving}. Thus previous work has approached terminology translation predominantly as a problem of making sure that the decoder's output contains \textit{lexically} and \textit{morphologically} pre-specified target language terms. While useful in some cases and some languages, such approaches come short of addressing terminology translation into morphologically complex languages where each word can have many morphological surface forms.

For terminology translation to be viable for translation into morphologically complex languages, terminology constraints have to be \textit{soft}. That is, terminology translation has to account for various natural language phenomena, which cause words to have more than one manifestation of their root morphemes. Multiple root morphemes complicate the application of \textit{hard} constraint methods, such as constrained-decoding \cite{hokamp-liu-2017-lexically}. That is because even after the terminology constraint is striped from the morphemes that encode all grammatical information, the remaining root morphemes still can be too restrictive to be used as hard constraints because, for many words, there can be more than one root morpheme possible. An illustrative example is the consonant mutation in the Latvian noun \textit{vācietis} (``the German'') which undergoes the mutation t$\rightarrow$š, thus yielding two variants of its root morpheme \textit{vācieš-} and \textit{vāciet-} \cite{bergmanis2020methods}. If either of the forms is used as a hard constraint for constrained decoding, the other one is excluded from appearing in the sentence's translation.

We propose a necessary modification for the method introduced by \newcite{dinu-etal-2019-training}, which allows training neural machine translation (NMT) systems that are capable of applying terminology  constraints: instead of annotating source-side terminology with their exact target language translations, we annotate randomly selected source language words with their target language lemmas. First of all, preparing training data in such a way relaxes the requirement for access to bilingual terminology resources at the training time. Second, we show that the model trained on such data does not learn to simply \textit{copy} inline annotations as in the case of  \newcite{dinu-etal-2019-training}, but learns \textit{copy-and-inflect} behaviour instead, thus addressing the need for \textit{soft} terminology constraints.

Our results show that the proposed approach not only relaxes the requirement for apriori specified target language forms but also yields substantial improvements over the previous work \cite{dinu-etal-2019-training} when tested on the morphologically complex Baltic and Uralic languages.

\begin{table}[]
\centering
\small
\begin{tabular}{ll}
\toprule
\textbf{EN Src.:} & \begin{tabular}[c]{@{}l@{}}faulty engine or in transmission{[}..{]}\end{tabular}  \\
\textbf{LV Trg.:}& atteice dzinējā vai transmisijas {[}..{]}         \\ 
\midrule
\textbf{ETA:} &
  \begin{tabular}[c]{@{}l@{}}faulty\textbar w engine\textbar s dzinēj\textbf{ā}\textbar t  or\textbar w \\ transmission\textbar s transmisij\textbf{as}\textbar t {[}..{]}\end{tabular} \\ 
  \midrule
\textbf{TLA:} &
  \begin{tabular}[c]{@{}l@{}}faulty\textbar w  engine\textbar s dzinēj\textbf{s}\textbar t  or\textbar w \\ transmission\textbar s transmisij\textbf{a}\textbar t {[}..{]}\end{tabular} \\
  \bottomrule
\end{tabular}
\caption{Examples of differences in input data in ETA \cite{dinu-etal-2019-training} and TLA (this work). Differences of inline annotations are marked in bold. \textbar w, \textbar s, \textbar t denote the values of the additional input stream and stand for regular words, source language annotated words, target language annotations respectively.    \label{table:input_example}}
\end{table}

\section{Method: Target Lemma Annotations}

To train NMT systems that allow applying terminology constraints \newcite{dinu-etal-2019-training} prepare training data by amending source language terms with their exact target annotations (\textbf{ETA}). To inform the NMT model about the nature of each token (i.e., whether it is a source language term, its target language translation or a regular source language word), the authors use an additional input stream---source-side factors \cite{sennrich-haddow-2016-linguistic}. Their method, however, is limited to cases in which the provided annotation matches the required target form and can be copied verbatim, thus performing poorly in cases where the surface forms of terms in the target language differ from those used to annotate source language sentences \cite{dinu-etal-2019-training}. This constitutes a problem for the method's practical applicability in real-life scenarios. In this work, we propose two changes to the approach of \newcite{dinu-etal-2019-training}. \textbf{First}, when preparing training data, instead of using terms found in either IATE\footnote{\url{https://iate.europa.eu}} or Wiktionary
%\footnote{\url{https://www.wiktionary.org/}} 
as done by \newcite{dinu-etal-2019-training}, we annotate random source language words. This relaxes the requirement for curated bilingual dictionaries for training data preparation. \textbf{Second}, rather than providing exactly those target language forms that are used in the target sentence, we use target lemma annotations (\textbf{TLA}) instead (see Table~\ref{table:input_example} for examples). We hypothesise that in order to benefit from such annotations, the NMT model will have to learn \textit{copy-and-inflect} behaviour instead of simple \textit{copying} as proposed by \newcite{dinu-etal-2019-training}. 

Our work is similar to work by \newcite{exel-etal-2020-terminology} in which authors also aim to achieve \textit{copy-and-inflect} behaviour. However, authors limit their annotations to only those terms for which their base forms differ by no more than two characters from the forms required in the target language sentence. Thus wordforms undergoing longer affix change or inflections accompanied by such linguistic phenomena as consonant mutation, consonant gradation or other stem change are never included in training data. 
\begin{figure*}
\centering

\includegraphics{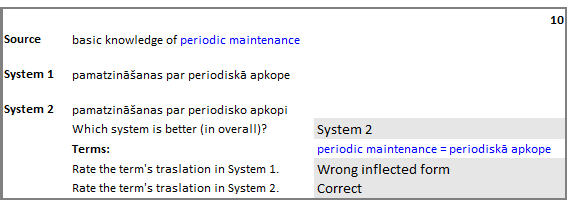}
\caption{Example of forms used in human evaluation.}
\label{fig:human_eval}
\end{figure*}

\begin{table}[]
\centering
\small
\begin{tabular}{lccc}
\toprule
                  & \textbf{Train}                     & \multicolumn{2}{c}{\textbf{Test}}                   \\
 &  & \begin{tabular}[c]{@{}l@{}}ATS\end{tabular} & \begin{tabular}[c]{@{}l@{}}WMT17+IATE\end{tabular} \\
 \midrule
EN-DE     & 27.6M &   768                   & 581      \\
EN-ET     & 2.4M  &  768 & -                    \\
EN-LV     & 22.6M &  768                    & -                    \\
EN-LT     & 22.1M &  768                    & -                    \\
\bottomrule
\end{tabular}
\caption{Training and evaluation data sizes in numbers of sentences. WMT2017 + IATE stands for the English-German test set from the news translation task of WMT2017 which is annotated with terminology from the IATE terminology database.}
\label{table:data-sizes}
\end{table}

\section{Experimental Setup}
\paragraph{Languages and Data.}
As our focus is on morphologically complex languages, in our experiments we translate from  English into Latvian and Lithuanian (Baltic branch of the Indo-European language family) as well as Estonian (Finnic branch of the Uralic language family). For comparability with the previous work, we also use English-German (Germanic branch of the Indo-European language family). For all language pairs, we use all data that is available in the \href{https://www.tilde.com/products-and-services/data-library}{Tilde Data Libarary} with an exception for English-Estonian for which we use data from WMT~2018. The size of the parallel corpora after pre-processing using the Tilde MT platform \cite{pinnis-etal-2018-tilde} and filtering tools \cite{pinnis2018tilde} is given in Table~\ref{table:data-sizes}.

To prepare data with TLA, we first lemmatise and part-of-speech (POS) tag the target language side of parallel corpora. For lemmatisation and POS tagging, we use pre-trained Stanza\footnote{\url{https://github.com/stanfordnlp/stanza}} \cite{qi2020stanza} models. 
We then use \textit{fast\_align}\footnote{\url{https://github.com/clab/fast_align}} \cite{dyer2013simple} to learn word alignments between the target language lemmas and source language inflected words. We only annotate verbs or nouns. To generate sentences with varying proportions of annotated and unannotated words, we first generate a sentence level annotation threshold uniformly at random from the interval $[0.6,1.0)$. Similarly, for each word in the source language sentence, we generate another number uniformly at random from the interval $[0.0,1.0)$. If the latter is larger than the sentence level annotation threshold, we annotate the respective word with its target language lemma. We use the original training data and annotated data with a proportion of 1:1. We follow \newcite{dinu-etal-2019-training} to prepare ETA and replicate their results.

For validation during training, we use development sets from the WMT news translation shared tasks. For EN-ET and EN-DE, we used the data from WMT~2018, for EN-LV -- WMT~2017, and for EN-LT -- WMT~2019.%\footnote{\url{http://www.statmt.org/wmt17|18|19}}.

\paragraph{MT Model and Training.}
For the most part, we use the default configuration of the Transformer \cite{vaswani2017attention} NMT model implementation of the Sockeye NMT toolkit \cite{Hieber2017Sockeye}. The exception is the use of source-side factors \cite{sennrich-haddow-2016-linguistic} with the dimensionality of 8 for systems using inline target lemma annotations. We train all models using early stopping with the patience of 10 based on their development set perplexity \cite{prechelt1998early}.
\begin{table*}[]
\centering
\small
\begin{tabular}{lll|ll|ll|ll|ll}
\toprule
 &
  \multicolumn{2}{c|}{\textbf{IATE}} &
  \multicolumn{8}{c}{\textbf{Automotive Test Suite}} \\ 
 &
  \multicolumn{2}{c|}{EN-DE}  & \multicolumn{2}{c}{EN-DE} &  \multicolumn{2}{c}{EN-ET} &  \multicolumn{2}{c}{EN-LV} &   \multicolumn{2}{c}{EN-LT} 
   \\ \midrule
 &
  BLEU &   Acc. &   BLEU &   Acc. &   BLEU &   Acc. &   BLEU &   Acc. &   BLEU &   Acc. \\ 
  \midrule
\textbf{Baseline} &   29.7 &   81.7 &    26.5 &   46.2 &  19.6 &   46.7 &  30.6 &  62.2 &  25.3 &  51.2 \\
\textbf{CD} & 28.5   &\textbf{99.7}  & 22.9 &  \textbf{99.7} & 14.9 & \textbf{98.0} & 23.5  & \textbf{99.3}  & 18.1 & \textbf{98.9} \\
\textbf{ETA} &   \textbf{29.9} &  96.2 &  $33.2^\dagger$ &  94.0 &  17.8 &  92.4 &   27.4 &   93.4 &  28.8$^\dagger$ &  89.7 \\ 
\midrule
\textbf{TLA}  &   29.5 &   96.5 &  \textbf{33.5}$^\dagger$ &  94.0 &  \textbf{21.0}$^{\dagger \ddagger}$ &  87.2 &   \textbf{35.0$^{\dagger \ddagger}$} &   92.0 &   \textbf{30.1}$^{\dagger \ddagger}$ &   90.3 \\
\bottomrule
\end{tabular}

\caption{Results of automatic evaluation metrics BLEU and term translation accuracy (Acc.). The numerically highest score in each column is given in bold; $^\dagger$ and $^\ddagger$ indicate statistically significant
\ul{improvements of BLEU} over Baseline and ETA respectively (all $p < 0.05$) .}
\label{table:results}
\end{table*}
\begin{table*}[h]
\small
\centering
\begin{tabular}{cc}
\begin{tabular}{lc|cc|c|c}
\toprule
         & \textbf{Correct} & \textbf{Wrong lexeme }& \textbf{Wrong inflect.  }  &     \textbf{Other}  &   \textbf{ $\kappa_\mathsf{free}$ }       \\
         
\midrule
\textbf{Basel.} & 55.1                     & 42.9      & 1.4          & 0.7  & 0.95       \\
\textbf{ETA}      & 45.2                     & 7.9      & 44.9          & 2.0  & 0.87       \\
\textbf{TLA}      & 92.9                     & 5.1      & 1.4          & 0.7  & 0.98      \\
\bottomrule
\end{tabular}
&
\begin{tabular}{ccc|c}
\toprule
\textbf{Baseline} & \textbf{Equal} & \textbf{TLA}  & \textbf{$\kappa_\mathsf{free}$ } \\
3.0      & 58.0  & 39.0 & 0.65      \\
\midrule
\textbf{ETA}      & \textbf{Equal} & \textbf{TLA}  & \textbf{$\kappa_\mathsf{free}$}\\

3.0     & 36.0  & 61.0 & 0.81     \\
\bottomrule
\end{tabular}
\end{tabular}
\caption{Results of human evaluation: term (on the left) and sentence (on the right) translation quality judgements in \%.  Sentence comparison is \ul{pairwise} contrasting TLA vs Baseline and TLA vs ETA. $\kappa$-free: inter-annotator agreement according to free marginal kappa \cite{randolph2005free}.}
\label{tab:human-eval}
\end{table*}

\paragraph{Evaluation Methods and Data.}
In previous work, methods were tested on general domain data\footnote{\url{https://github.com/mtresearcher/terminology_dataset}} annotated with exact surface forms of general-domain words from IATE and Wiktionary. Although data constructed in such a way is not only artificial but also gives an oversimplified view on terminology translation, we do use the data from IATE to validate our re-implementation of the method from \newcite{dinu-etal-2019-training}. Other than that, we test on the Automotive Test Suite\footnote{\url{https://github.com/tilde-nlp/terminology_translation}} (ATS): a data set containing translations of the same 768 sentences in English, Estonian, German, Latvian, and Lithuanian. ATS contains about 1.1k term occurrences from a glossary prepared by professional translators. When annotating terms in the source text, we use only the dictionary forms of term translations, since in practical applications having access to the correct inflections (surface forms) is unrealistic.

We compare our work with an NMT system without means for terminology integration (\textbf{Baseline}) and the previous work by \newcite{dinu-etal-2019-training} (\textbf{ETA}). Although our preliminary experiments with constrained decoding \cite{post2018fast} (\textbf{CD}) confirmed the findings by \newcite{dinu-etal-2019-training} that strict enforcement of constraints leads to lower-than-baseline quality, we nevertheless include them for completeness sake. 

Similarly to the previous work, we use two automatic means for evaluation: BLEU \cite{papineni2002bleu} and lemmatised term exact match accuracy. We use BLEU as an extrinsic evaluation metric as we expect that, when successful, the methods for terminology translation should yield substantial overall translation quality improvements due to correctly translated domain-specific terms. For significance testing, we use pairwise bootstrap re-sampling \cite{koehn2004statistical}. We use lemmatised term exact match accuracy as an intrinsic metric because it directly measures the adequacy of terminology translation (i.e., whether or not the correct lexeme appears in the target sentence).

We are aware that the automatic evaluation methods are merely an approximation of translation quality. For example, we use lemmatised term exact match accuracy to measure term use in target language translations; however, it does not capture whether the term is inflected correctly. Thus human evaluation is in place.
We use the EN-LV language pair to compare TLA against baseline and ETA. We use a 100 sentences large randomly selected ATS subset that contains 147 terms of the original test suite. We employ four professional translators and Latvian native speakers to compare each system's translations according to their overall translation quality and judge individual term translation quality. 
Specifically, given the original sentence and its two translations (in a randomised order), raters are asked to answer ``which system's translation is better overall?''. Raters are also given a list of the terms being evaluated and their reference translations (from the term collection) and are asked to classify translations as either ``Correct'', ``Wrong lexeme'', ``Wrong inflection'', or ``Other''. Figure~\ref{fig:human_eval} gives an example of the forms presented to raters during the human evaluation of term and overall translation quality. We report inter-annotator agreement using free marginal kappa, $\kappa_\mathsf{free}$ \cite{randolph2005free}.
\vspace{-6px}

\section{Results}\vspace{-3px}
\paragraph{Automatic Evaluation.}
We first validate our re-implementation of ETA by testing on the English-German WMT~2017 test set annotated with terms from IATE as used by \newcite{dinu-etal-2019-training}. Results (see columns 2 and 3 of Table~\ref{table:results}) are similar to those of the previous work: on this data set, ETA yields minor translation quality improvements over the baseline (+0.2 BLEU) and considerable improvement (+14.5\%) in term translation accuracy. 

When evaluated on the ATS, systems using TLA always yield results that are better than the baseline both in terms of BLEU scores (+1.4--7 BLEU) and term translation accuracy (29.8\%--47.8\%) (see columns 4-11 of Table~\ref{table:results}). Results also show that when compared to ETA, systems integrating terminology using TLA achieve statistically significant improvements in terms of BLEU scores for three out of four languages-pairs. An exception is EN-DE, for which both systems, ETA and TLA, perform similarly. Analysing reference translations of the EN-DE language pair, we find that as many as 87\% of the German terms are used in their dictionary forms, which explains the comparable performance of systems trained using ETA and TLA on EN-DE.

Results also confirm the finding of the previous work by \newcite{dinu-etal-2019-training} and \newcite{exel-etal-2020-terminology}, that the strict enforcement of constraints by constrained decoding leads to lower-than-baseline BLEU scores on all data sets for all languages. BLEU scores are abysmal when translating into the morphologically complex languages as for these languages citation form seldom happens to be the form required in the target language sentence. This result further illustrates why terminology constraints have to be \textit{soft} when translating into morphologically complex languages.\vspace{-6px}

\paragraph{Human Evaluation.} 
Results of human evaluation of EN-LV systems are summarised in Table~\ref{tab:human-eval}.
First, we note that on this dataset, the baseline system translates terms correctly 55\% of the time, yet it makes mistakes by choosing the wrong lexeme for most of the other cases (Table~\ref{tab:human-eval}, left). The system using ETA, on the other hand, has a much lower rate of correctly translated terms -- 45\%, which roughly corresponds to the proportion of Latvian terms in the reference translations that are used in their dictionary forms (47\%). The remaining cases are mistranslated by choosing the wrong inflected form. The system using TLA, in comparison, does very well as it gets terminology translations right 93\% of the time. 
Examining the cases where terms had been mistranslated by choosing the wrong lexeme, we find that most of these cases are multi-word terms with some other word inserted between their constituent parts. 
The high $\kappa$-$\mathsf{free}$ values indicate almost perfect inter-annotator agreement suggesting that the task of term translation quality evaluation has been easy and results are reliable.  

The overall sentence translation quality judgements (Table~\ref{tab:human-eval}, right) also favour translations produced by the system using TLA deeming it better than or on par with the baseline system and system using ETA 97\% of the time. The system using TLA is strictly favoured over its ETA counterpart for 61\% of the translations. Again, annotators have reached an almost perfect agreement ($\kappa_\mathsf{free}=0.81$) when comparing the systems using TLA and ETA, suggesting that the task has been easy. 
These results clearly show that at least for the EN-LV language pair and the test set considered here, systems using TLA improve term translation quality by correctly choosing adequate translations and morpho-syntactically appropriate inflections. \vspace{-5px}

\paragraph{Productivity of NMT models.} Terminology translation frequently involves the translation of niche lexemes with rare or even unseen inflections. Thus the model's ability to generate novel wordforms is critical for high-quality translations. 
To verify if our NMT models are lexically and morphologically productive, we analysed Latvian translations of ATS produced by the system using TLA and looked for wordforms that are not present in either source or target language side of the training data. We found 72 such wordforms. Of those 45 or 62.5\% were valid wordforms that were not present in training data, of which 28 were novel inflections related to ATS terminology use, while the remaining 17 where novel forms of general words. We interpret this as \textit{some} evidence that the NMT model, when needed, generates novel wordforms.
The remaining 27 or 37.5\% were not valid, albeit sometimes plausible, Latvian language words, common types of errors being literal translations and transliterations of English words as well as words that would have been correct, if not for errors with consonant mutation. \vspace{-5px}

\section{Conclusions}\vspace{-3px}
We proposed TLA---a flexible and easy-to-implement method for terminology integration in NMT. Using TLA does not require access to bilingual terminology resources at system training time as it annotates ordinary words with lemmas of their target language translations. This simplifies data preparation greatly and also relaxes the requirement for apriori specified target language forms during the translation, making our method practically viable for terminology translation in real-life scenarios. Results from experiments on three morphologically complex languages demonstrated substantial and systematic improvements over the baseline NMT systems without means for terminology integration and the previous work both in terms of automatic and human evaluation judging term and overall translation quality.   \vspace{-4px}

\section*{Acknowledgements}\vspace{-6px}
This research has been supported by the ICT Competence Centre (www.itkc.lv) within the project “2.2. Adaptive Multimodal Neural Machine Translation” of EU Structural funds, ID no 1.2.1.1/18/A/003.

\bibliography{eacl2021}
\bibliographystyle{acl_natbib}
\end{document}